\pdfoutput=1

\documentclass[11pt]{article}
\usepackage[usenames,dvipsnames]{color}
\usepackage[]{acl}

\usepackage{gb4e}
\noautomath
\usepackage{paralist}
\usepackage{mathtools}
\usepackage[inline]{enumitem}
\usepackage{times}
\usepackage{latexsym}
\usepackage{amsmath}
\usepackage[capitalize]{cleveref}
\usepackage{amssymb}
\usepackage{multirow,multicol}
\usepackage{graphicx}

\usepackage{array}
\usepackage{bm}
\usepackage{soul}
\usepackage{booktabs}
\usepackage[colorinlistoftodos]{todonotes}
\usepackage{enumitem}
\usepackage{amsthm}
\usepackage{caption}
\usepackage{subcaption}
\usepackage{tipa}
\usepackage{stackengine}
\usepackage{rotating}
\usepackage{tabularx}
\usepackage[T1]{fontenc}
\usepackage[utf8]{inputenc}
\usepackage{multirow}

\newcommand{\declarelogo}[0]{\includegraphics[height=.02\textwidth]{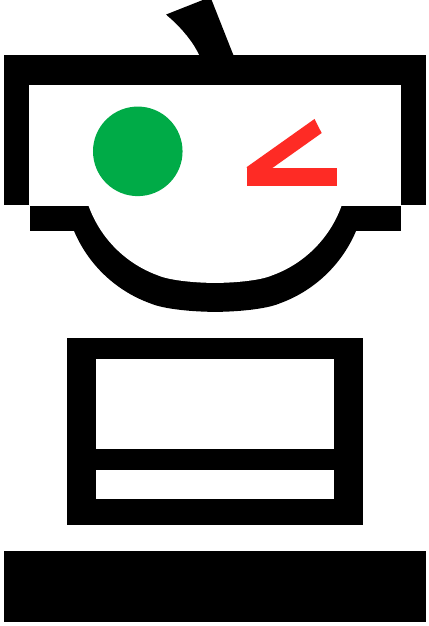}}
\newcommand{\umichlogo}[0]{\includegraphics[height=.012\textwidth]{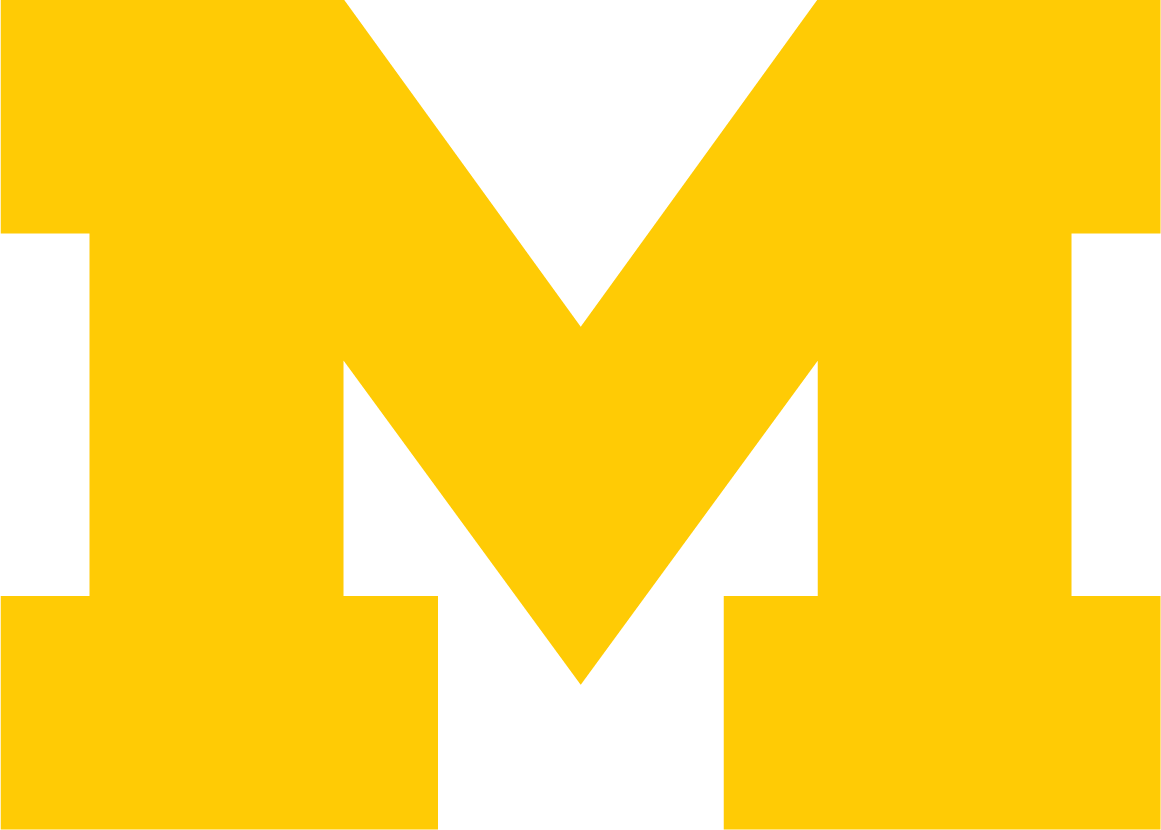}}

\newcommand{\PreserveBackslash}[1]{\let\temp=\\#1\let\\=\temp}
\newcolumntype{C}[1]{>{\PreserveBackslash\centering}p{#1}}
\newcolumntype{R}[1]{>{\PreserveBackslash\raggedleft}p{#1}}
\newcolumntype{L}[1]{>{\PreserveBackslash\raggedright}p{#1}}

\newcommand\code[1]{\texttt{#1}}

\crefformat{section}{\S#2#1#3} %
\crefformat{subsection}{\S#2#1#3}
\crefformat{subsubsection}{\S#2#1#3}
\newcommand{\model}{{\code{TEAM}}}
\newcommand{\scoremodel}{{\code{Score}}}

\title{An Empirical Report of Multi-Choice Question Answering \\ with Binary Classification}

\title{Approaching Multi-Choice Question Answering using Binary Classification: An Empirical Report and Analysis}

\title{Two is Better than Many? Binary Classification as an\\ Effective Approach to Multi-Choice Question Answering} 

\author{First Author \\
  Affiliation / Address line 1 \\
  \texttt{email@domain} \\\And
  Second Author \\
  Affiliation / Address line 1 \\
  \texttt{email@domain} \\}

\author{Deepanway Ghosal$^{\declarelogo}$ \hspace{2mm}
Navonil Majumder$^{\declarelogo}$ \hspace{2mm}\\
\textbf{Rada Mihalcea$^{\umichlogo}$ \hspace{2mm}
Soujanya Poria$^{\declarelogo}$}\\
  $^{\declarelogo}$ DeCLaRe Lab, Singapore University of Technology and Design, Singapore\\
  $^{\umichlogo}$ University of Michigan, USA\\
  \texttt{\{deepanway\_ghosal@mymail.,navonil\_majumder@,sporia@\}sutd.edu.sg}\\ \texttt{mihalcea@umich.edu}\\
  }
  
\begin{document}
\maketitle
\begin{abstract}
We propose a simple refactoring of multi-choice question answering (MCQA) tasks as a series of binary classifications. The MCQA task is generally performed by scoring each (question, answer) pair normalized over all the pairs, and then selecting the answer from the pair that yield the highest score. For \textit{n} answer choices, this is equivalent to an \textit{n}-class classification setup where only one class (true answer) is correct. We instead show that classifying (question, true answer) as positive instances and (question, false answer) as negative instances is significantly more effective across various models and datasets. We show the efficacy of our proposed approach in different tasks -- abductive reasoning, commonsense question answering, science question answering, and sentence completion. Our DeBERTa binary classification model reaches the top or close to the top performance on public leaderboards for these tasks. The source code of the proposed approach is available at \url{https://github.com/declare-lab/TEAM}.
\end{abstract}

\section{Introduction}

Starting with the early Text Retrieval Conference (TREC) community-wide evaluations of textual question answering  \cite{voorhees1999trec}, all the way to the recent work on multimodal question answering \cite{Lei18,Tapaswi16,Jang17,lifeqa} and commonsense question answering \cite{sap2019social,talmor2019commonsenseqa}, the task has become a staple of the natural language processing research community. One of the major challenges encountered in question answering is the evaluation, which often requires human input to evaluate the textual answers thoroughly. Because of this, the alternative that has been proposed is that of {\it multi-choice question answering}, where the correct answer is provided together with other incorrect answers. The task is thus transformed into that of answer classification, where a system has to select one answer from the choices provided. While there are drawbacks associated with this evaluation metric, it has been widely adopted because of its benefit of providing a clear evaluation methodology. 

\looseness=-1 In this paper, we reformulate the task of multi-choice question answering as a binary classification task and show that this re-framing leads to significant performance improvements on several datasets. Importantly, this formulation brings flexibility to the overall question-answering setup, as it reduces the dependence on the up-front availability of multiple candidate answers. Using our method -- \model{} (\textbf{T}wo is b\textbf{E}tter th\textbf{A}n \textbf{M}any), candidate answers can be produced and evaluated for correctness on the fly, and thus the answer classification component can be also used in conjunction with more natural settings that use open-ended answer generation \cite{castro-etal-2022-fiber,sadhu-etal-2021-video}.

\section{Methodology}
\label{sec:method}
Let $q$ be a question for which multiple answer choices $\mathcal{A} = \{a_1, \dots, a_n\}$ are given. Optionally, there is some context $c$ which could be helpful for answering the question. The objective is to select the correct answer $a_{k}$ from the answer set $\mathcal{A}$. 

For some of the datasets used in the paper, the question $q$ is not provided, and the answer is based only on the context $c$. For example, SWAG and HellaSwag are two such datasets where the task is to choose the best possible ending for sentence completion, as shown in \cref{tab:datasets}. In this case, the question $q$ can be assumed as implicit: \textit{What is the best possible ending for the context?} The sentence to be completed is considered as the context $c$.

We discuss how the MCQA task is generally performed using transformer language models in \cref{sec:score}. We denote this approach as \textbf{Score-based Method} or \scoremodel{} method . We then discuss our proposed \textbf{Binary Classification-based Method}, \model{} in \cref{sec:cls}. 

\subsection{Score-based Method (\scoremodel{})}
\label{sec:score}
We use the notation introduced earlier in \cref{sec:method}. Given question $q$, optional context $c$, and the answer choices $\mathcal{A} = \{a_1, a_2, \dots, a_n\}$, $n$ different input sequences are constructed each containing the concatenation of the question $q$, context $c$, and one possible answer choice $a_i$. The sequences are independently encoded through a pre-trained transformer language model such as RoBERTa~\cite{liu2019roberta} or DeBERTa~\cite{he2021debertav3}. A score $s_i$ is predicted for each input sequence which is then normalized with a softmax layer across the $n$ outputs to obtain score $q_i$.

The cross-entropy loss is used to train the encoder model. Assuming the answer $a_k$ is correct, the loss can be obtained as follows:
\begin{equation}
\setlength\abovedisplayskip{7pt}
\setlength\belowdisplayskip{7pt}
\begin{split}
\mathcal{L} & = - \sum_{i=1}^{n} p_i log(q_i) = -log(q_k) \\
\end{split}
\end{equation}
where $p_i$ are considered as the class labels. The class $p_k$ corresponding to the gold answer $a_k$ is valued as 1, and all other classes are valued as 0. The loss is equivalent to the cross-entropy loss in a $n$-class classification setup. The normalization of the scores using the softmax layer to obtain a distribution over the answer choices is also analogous to the probability distribution over the different classes in the multi-class classification setup. 

The choice providing the highest score is the predicted answer during inference. The \scoremodel{} method was used for the SWAG task in BERT~\cite{devlin-etal-2019-bert}, StoryCloze task in GPT~\cite{radford2018improving} and has been used for all MCQA tasks in the huggingface transformers\footnote{\url{https://github.com/huggingface/transformers}} framework.

\subsection{Classification-based Method (\model{})}
\label{sec:cls}
For our proposed classification-based method, we first extend the pre-trained language model by adding a classification head with two nodes. The values of these two nodes will denote the unnormalized scores for the negative and positive classes in our classification setup.

Now, similar to the previous \scoremodel{} method, we first construct $n$ different input sequences by concatenating the question $q$, the optional context $c$, and each possible answer choice $a_i$. We then obtain the unnormalized negative and positive scores $s_i^-$ and $s_i^+$ for each sequence by independently encoding them through the modified language model. We normalize each pair of scores through a softmax layer to obtain probabilities of negative and positive classes: $q_i^-$ and $q_i^+$, respectively.

We consider the sequence corresponding to the gold answer $a_k$ as positive, and all the other sequences as negative. Therefore, the loss function takes the following form:
\begin{equation}
\setlength\abovedisplayskip{5pt}
\setlength\belowdisplayskip{5pt}
\begin{split}
\mathcal{L} & = - \sum_{i=1}^{n} (p_i^+ log(q_i^+) + p_i^{-} log(q_i^-))\\
& = - log(q_k^+) - \sum_{i=1, i\neq k}^{n} log(q_i^-)
\end{split}
\label{eq:cls}
\end{equation}
where $p_i^+$ and $p_i^-$ are considered as the class labels. As $a_k$ is the gold answer, we use $p_k^+ = 1$, $p_k^- = 0$ and $p_i^+ = 0$, $p_i^- = 1$, when $i \neq k$.

Although \cref{eq:cls} is a suitable loss function for single correct answer cases, it can be easily extended for instances or datasets with multiple correct answers. This can be done by changing the class labels $p_i^+$ and $p_i^-$ to positive and negative appropriately for the additional correct answers.

During inference, we choose the answer with the highest positive class probability as the predicted answer. 
We will show later in \cref{sec:results} that the \model{} method generally outperforms the \scoremodel{} method across several datasets for the same choice of transformer models.

\section{Experimental Datasets}
We experiment with the following datasets:

\noindent \textbf{Abductive NLI}~\cite{bhagavatula2020abductive}. Given two observations $o_1$ and $o_2$ (considered as context $c$), the goal is to select the more plausible intermediate event among hypotheses $h_1$ and $h_2$. 
We use the sequences $\{o_1, h_1, o_2\}$ and $\{o_1, h_2, o_2\}$ as input for both the \scoremodel{} and \model{} method. Assuming $h_1$ is the gold answer, we classify $\{o_1, h_1, o_2\}$ as positive; $\{o_1, h_2, o_2\}$ as negative.

\vspace{0.9mm}
\noindent \textbf{CommonsenseQA}~\cite{talmor2019commonsenseqa} or CQA is a dataset for commonsense QA based on knowledge encoded in ConceptNet~\cite{speer2017conceptnet}.
Given a question, there are five possible choices
, among which only one is correct. 
We do not use any additional knowledge or context for this task. 

\vspace{0.9mm}
\noindent \textbf{CommonsenseQA 2.0}~\cite{talmor2021commonsenseqa} or CQA2 is a recent challenging QA dataset collected with a model-in-the-loop approach. 
The dataset contains commonsense questions from various reasoning categories 
with either \textit{yes} or \textit{no} answer.

\vspace{0.9mm}
\noindent \textbf{QASC}~\cite{khot2020qasc} or Question Answering via Sentence Composition task requires fact retrieval from a large corpus and composing them to answer a multi-choice science question.
Each question $q$ has eight choices, among which one is correct. We use the question and choices without any retrieved facts for this task. We evaluate another task setup \textbf{QASC-IR} (information retrieval) where we use two-step IR retrieved facts as in ~\citet{khot2020qasc} as additional context $c$.

\newcommand{\emerald}[1]{\textcolor{Emerald}{#1}}

\begin{table}[t]
\small
\centering
\resizebox{\linewidth}{!}{
\begin{tabular}{lL{9cm}}
\toprule
\textbf{Dataset} & \textbf{Instance} \\
\midrule

\multirow{2}{*}{CQA} & 
\textbf{Question:} Where on a river can you hold a cup upright to catch water on a sunny day? \\
\cmidrule{2-2}
& \textbf{Choice 1:} \color{Emerald}{Waterfall} \hspace{2mm} \color{black}\textbf{Choice 2:} Bridge \hspace{2mm} $\dots$ \hspace{2mm}  \textbf{Choice 5:} Mountain \\
\midrule

\multirow{4}{*}{QASC} & 
\textbf{Question:} Differential heating of air can be harnessed for
what? \\
\cmidrule{2-2}
& \textbf{Choice 1:} \color{Emerald}{electricity production} \hspace{2mm} \color{black}\textbf{Choice 2:} running and lifting \hspace{2mm} \\
& \textbf{Choice 3:} animal survival  $\, \ \ \ \dots \quad $ \textbf{Choice 8:} reducing acid rain \\
\midrule

\multirow{5}{*}{SWAG} & 
\textbf{Partial Event:} On stage, a woman takes a seat at the piano. She \\
\cmidrule{2-2}
& \textbf{Ending 1:} sits on a bench as her sister plays with the doll. \\
& $\dots$ \\
& \textbf{Ending 4:}  \color{Emerald}{nervously sets her fingers on the keys.} \\
\midrule

\multirow{6}{*}{PIQA} & 
\textbf{Goal:} To separate egg whites from the yolk using a water bottle, you should \\
\cmidrule{2-2}
& \textbf{Solution 1:} \color{Emerald}{Squeeze the water bottle and press it against the yolk. Release, which creates suction and lifts the yolk.} \\
& \textbf{Solution 2:} Place the water bottle and press it against the yolk. Keep pushing, which creates suction and lifts the yolk.\\
\bottomrule

\end{tabular}
}
\caption{\footnotesize Illustration of some of the datasets used in this work. The answers highlighted in \emerald{green} are the correct answers. CQA: Commonsense QA, PIQA: Physical IQA.}
\label{tab:datasets}
\end{table}

\begin{table*}[h!]
\centering
\resizebox{\linewidth}{!}{
\begin{tabular}{lccccccccccccc}
\toprule
\multirow{2}{*}{\textbf{Model}} & \multirow{2}{*}{\textbf{Method}} & \multirow{2}{*}{\textbf{ANLI}} & \multirow{2}{*}{\textbf{CQA}} & \multirow{2}{*}{\textbf{CQA2}} & \multirow{2}{*}{\textbf{QASC}} & \multirow{2}{*}{\textbf{QASC-IR}} & \multirow{2}{*}{\textbf{SWAG}} & \multirow{2}{*}{\textbf{H-SWAG}} & \multirow{2}{*}{\textbf{SIQA}} & \multirow{2}{*}{\textbf{PIQA}} & \multirow{2}{*}{\textbf{CosmosQA}} & \multicolumn{2}{c}{\textbf{CICERO$^{*}$}} \\
& & & & & & & & & & & & v1 & v2 \\
\midrule
\multirow{2}{*}{RoBERTa Large} & \scoremodel{} & 85.25 & 73.63 & 54.76 & 53.46 & 77.21 & 89.23 & 83.89 & \textbf{78.15} & \textbf{78.89} & \textbf{80.44} & \textbf{80.33} & 85.25 \\
& \model{} & \textbf{87.47} & \textbf{75.32} & \textbf{55.83} & \textbf{57.24} & \textbf{80.35} & \textbf{89.49} & \textbf{84.52} & 76.49 & 76.71 & 80.37 & 77.54 & \textbf{86.53} \\
\midrule
\multirow{2}{*}{DeBERTa Large} & \scoremodel{} & 89.75 & \textbf{83.75} & 66.63 & 74.41 & 89.31 & 93.14 & 94.67 & \textbf{80.82} & \textbf{87.81} & 86.13 & \textbf{86.60} & 89.06 \\
& \model{} & \textbf{92.23} & 83.34 & \textbf{69.57} & \textbf{75.33} & \textbf{91.09} & \textbf{93.27} & \textbf{95.47} & 80.27 & 86.07 & \textbf{86.35} & 84.48 & \textbf{90.59} \\
\bottomrule
\end{tabular}
}
\caption{\footnotesize{Accuracy on the validation split of the datasets. 
All numbers are the average of five runs with different seeds. 
}}
\label{tab:val-results}

\vspace{5mm}

\centering
\resizebox{\linewidth}{!}{
\begin{tabular}{lccccccccccccc}
\toprule
\multirow{2}{*}{\textbf{Model}} & \multirow{2}{*}{\textbf{Method}} & \multirow{2}{*}{\textbf{ANLI}} & \multirow{2}{*}{\textbf{CQA2}} & \multirow{2}{*}{\textbf{QASC}} & \multirow{2}{*}{\textbf{QASC-IR}} & \multirow{2}{*}{\textbf{SWAG}} & \multirow{2}{*}{\textbf{H-SWAG}} & \multirow{2}{*}{\textbf{SIQA}} & \multirow{2}{*}{\textbf{PIQA}} & \multirow{2}{*}{\textbf{CosmosQA}} & \multicolumn{2}{c}{\textbf{CICERO$^{*}$}} \\
& & & & & & & & & & & v1 & v2 \\
\midrule
\multirow{2}{*}{RoBERTa Large} & \scoremodel{} & \textcolor{purple}{83.91} & 55.44 & 46.52 & 73.26 & 88.97 & \textcolor{purple}{81.70} & \textcolor{purple}{\textbf{76.70}} & \textcolor{purple}{\textbf{79.40}} & 80.71 & \textbf{83.28} & 89.61 \\
& \model{} & \textbf{87.04} & \textbf{56.73} & \textbf{53.80} & \textbf{77.93} & \textbf{89.88 (7)}  & \textbf{83.63} & 75.96 & 74.55 & \textbf{80.84} & 79.94 & \textbf{89.81} \\
\midrule
\multirow{2}{*}{DeBERTa Large} & \scoremodel{} & 89.74 & 67.37 & 71.74 & 85.65 & 92.37 (2) & 94.72 (4) & \textbf{80.18} & \textbf{87.41 (4)} & 85.51 & \textbf{88.04} & 92.67 \\
& \model{} & \textbf{92.20 (1)} &  \textbf{68.38 (9)} & \textbf{74.35}  & \textbf{89.35 (3)} &  \textbf{94.12 (1)} & \textbf{95.57 (2)} & 79.89 & 85.90 (5) & \textbf{86.86 (5)} & 86.84 & \textbf{93.25} \\
\midrule
UnifiedQA 11B & - & - & - & \textbf{78.50} & \textbf{89.60} & - & - & 81.40 & 89.50 & - & - & - \\
UNICORN 11B & - & 87.30 & \textbf{70.20} & - & - & - & 93.90 & \textbf{83.20} & \textbf{90.10} & \textbf{91.80} & - & - \\
\bottomrule
\end{tabular}
}
\caption{\footnotesize{Accuracy on the test split of the datasets. Numbers on the parentheses indicate rank on the leaderboard (if in the top 10) at the time of submission to the leaderboard. Numbers in \textcolor{purple}{purple} indicate results for RoBERTa Large as reported in the UNICORN paper~\cite{lourie2021unicorn}. We do not report results for CommonsenseQA (CQA) test set as test labels are not publicly available and there is no automated submission leaderboard.}}
\label{tab:test-results}
\end{table*}

\vspace{0.9mm}
\noindent \textbf{SWAG, HellaSwag}~\cite{zellers-etal-2018-swag, zellers2019hellaswag} are two datasets for grounded commonsense inference, where the objective is to find the correct ending given a partial description of an event. 
We consider the partial description as the context $c$. The correct ending is to be chosen from a pool of four possible choices.

\vspace{0.9mm}
\noindent \textbf{Social IQA} (SIQA)~\cite{sap2019social} is a dataset for commonsense reasoning about social interactive situations. Given a question about a social situation context, the objective is to select the correct answer from three possible choices.

\vspace{0.9mm}
\noindent \textbf{Physical IQA} (PIQA)~\cite{bisk2020piqa} is designed to investigate physical knowledge of language models. 
The task is to select the correct solution for a goal from two given choices.

\vspace{0.9mm}
\noindent \textbf{CosmosQA}~\cite{huang2019cosmos} is a QA dataset for  commonsense-based reading comprehension. Given a question about a paragraph ($c$), the task is to select the correct answer among four choices.

\vspace{0.9mm}
\noindent \textbf{CICERO v1, v2}~\cite{ghosal2022cicero,shen2022multiview} are datasets for contextual commonsense reasoning in dialogues. Given the dialogue and a question about an utterance, the task is to choose the correct answer among multiple choices. We modify the original datasets to use them in a MCQA setup. More details are in the appendix.

\section{Results} \label{sec:results}
We use the RoBERTa Large~\cite{liu2019roberta} and DeBERTa Large~\cite{he2021debertav3} model to benchmark the \scoremodel{} and \model{} method across the experimental datasets. We report the accuracy for the validation set in \cref{tab:val-results} and accuracy of leaderboard submissions for the test set in \cref{tab:test-results}. We also report results for other QA systems such as UnifiedQA~\cite{khashabi-etal-2020-unifiedqa} and UNICORN~\cite{lourie2021unicorn} for the test set (wherever available) in \cref{tab:test-results}.

Our main finding is that the \model{} method improves over the \scoremodel{} method for most of the datasets except Social IQA, Physical IQA, and CICERO v1. We observe this result for both the RoBERTa and DeBERTa models. 

\vspace{1.65mm}
\noindent \textbf{Abductive Reasoning:} The improvement is consistently large for both validation and test set in the Abductive NLI (ANLI) dataset. The problem of intermediate hypothesis selection transforms into a problem of plausible story selection as we use the sequence $\{o_1, h, o_2\}$ as our input. In this formulation, the \model{} method is significantly better than the \scoremodel{} method for both RoBERTa and DeBERTa models.

\vspace{1.75mm}
\noindent \textbf{Science QA:} We also observe considerable improvements in the QASC dataset without and with the additional retrieved knowledge. The RoBERTa-\model{} model is 
more than 7\% better in the test set when retrieved knowledge is not used. The difference in performance is around 3\% and 4.5\% in the validation and test set when the retrieved knowledge is used. 
For DeBERTa, we observe the most significant improvement in the test results of the QASC-IR setting, where the \model{} method is 3.7\% better than the \scoremodel{} method.

\vspace{2.05mm}
\noindent \textbf{Commonsense QA and Sentence Ending Prediction:} The \model{} method is also better than the \scoremodel{} method for commonsense question-answering in CommonsenseQA and CommonsenseQA 2.0 across most settings. One notable instance is the 3\% superior score of the DeBERTa \model{} in the CommonsenseQA 2.0 validation set. We observe a similar trend in results for sentence-ending prediction in SWAG and HellaSwag. The improvement in performance for the \model{} method is between 0.85-1.9\% in the test set. We also notice improvements in the test set results for reading comprehension QA in CosmosQA.

\vspace{2.05mm}
\noindent \textbf{Dialogue Commonsense Reasoning:} We observe contrasting results in CICERO v1 and v2. The \scoremodel{} method outperforms the \model{} method by around 2-3\% in CICERO v1. However, the \model{} method is better in CICERO v2 for both RoBERTa and DeBERTa models. We analyze the results in more detail in \Cref{sec:analysis-answer-similarity}.

\vspace{2.05mm}
\noindent \textbf{Negative Results:} The \scoremodel{} method outperforms the \model{} method in Physical IQA (PIQA) and CICERO v1. These two datasets contain answer choices that are lexically close together and subtly different from each other (example in \cref{tab:datasets}). We analyze the results in more detail in \Cref{sec:analysis-answer-similarity}. The \scoremodel{} method is also the better performing method in SIQA, with small improvements over the \model{} method in DeBERTa and comparatively large improvements in RoBERTa. We surmise that the \scoremodel{} method is better because the dataset contains complex social commonsense scenarios, for which learning by directly comparing the options is more effective.

\vspace{2mm}
\noindent \textbf{State-of-the-Art Models and Leaderboard Submissions:}  We also report the results for UnifiedQA and UNICORN 11B models for the test set in \cref{tab:test-results}. We compare these results against our best-performing model: DeBERTa Large in classification setup (DeBERTa-\model{}). 
DeBERTa-\model{} maintains parity with UnifiedQA 11B in QASC-IR, despite being 36 times smaller. UNICORN 11B outperforms DeBERTa-\model{} by a large margin on SIQA, PIQA, and CosmosQA. It is an expected result as UNICORN is trained on multiple datasets for commonsense reasoning starting from the T5-11B checkpoint and then finetuned on each target dataset. DeBERTa-\model{} is, however, considerably better in Abductive NLI and HellaSwag. DeBERTa-\model{} also reached the top or close to the top of the leaderboard (at the time of submission to the leaderboard) in Abductive NLI, SWAG, HellaSwag, and QASC. 

\section{Analysis}
\subsection{How Does Similar Answer Choices Affect Performance?}
\label{sec:analysis-answer-similarity}

We analyze the similarity between the correct and incorrect choices to understand why the \model{} method is better than the \scoremodel{} method in most of the datasets and vice-versa in the others. We report the lexical similarity with BLEU~\cite{papineni2002bleu}, ROUGE-L~\cite{lin2004rouge}, and semantic similarity with \textit{all-mpnet-base-v2} sentence transformer ~\cite{reimers-2019-sentence-bert} in \Cref{tab:answer-similarity}. We also report the difference in performance between \model{} and \scoremodel{} models for RoBERTa and DeBERTa in the $\Delta$ columns.

The similarity measurements in \Cref{tab:answer-similarity} indicate that the datasets can be clearly segregated into two groups -- one with low to medium similarity, and the other with very high similarity. Interestingly, the $\Delta$ values are mostly positive for the low to medium similarity group, and all negatives for the high similarity group. We surmise that the difference between the very similar correct and incorrect choices are better captured through the softmax activation over the answers in the \scoremodel{} method. However, this aspect is not captured in the \model{} method, as sequences corresponding to the correct and incorrect choices are separately classified as positive or negative. Thus, the \scoremodel{} method is more effective when the answer choices are very similar, as in PIQA or CICERO v1.

\begin{table}[t!]
\centering
\resizebox{\linewidth}{!}{
\begin{tabular}{l|cccc|cc}
\toprule
\textbf{Dataset} & \textbf{BLEU1} & \textbf{BLEU4} & \textbf{ROUGE} & \textbf{Sem-Sim} & $\bf{\Delta_1}$ & $\bf{\Delta_2}$ \\
\toprule
ANLI & 21.84 & 7.81 & 24.61 & 46.02 & 2.22 & 2.48 \\
CQA & 1.48 & 0 & 1.31 & 30.75 & 1.69 & -0.41 \\
QASC & 3.15 & 0.95 & 2.08 & 25.71 & 3.14 & 1.78 \\
SWAG & 12.78 & 0.81 & 11.61 & 30.47 & 0.26 & 0.13 \\
H-SWAG & 18.55 & 1.18 & 16.14 & 46.95 & 0.63 & 0.80 \\
SIQA & 12.56 & 3.99 & 10.41 & 29.17 & -1.66 & -0.55 \\
CosmosQA & 32.37 & 13.31 & 24.66 & 35.29 & -0.07 & 0.22 \\
CICEROv2 & 30.00 & 7.50 & 33.85 & 44.23 & 1.28 & 1.53 \\
\midrule
PIQA & 81.97 & 72.77 & 74.01 & 82.50 & -2.18 & -1.74 \\
CICEROv1 &	73.17 & 53.96 & 74.98 & 74.12 & -2.79 & -2.12 \\
\bottomrule
\end{tabular}
}
\caption{
\footnotesize{Average similarity between correct and incorrect answer choices in the validation set for different datasets. Numbers are shown on a scale of 0-100. $\Delta_1$ and $\Delta_2$ indicate difference in performance between \model{} and \scoremodel{} methods for RoBERTa and DeBERTa in validation set.}
}
\label{tab:answer-similarity}
\end{table}

\subsection{How Accurate is the Binary Classifier?}
We evaluate how often input sequences corresponding to correct and incorrect answers are predicted accurately with DeBERTA-\model{} binary classification model in \Cref{tab:classifier-analysis}. The binary classifier model is more likely to predict all answers as negative than all answers as positive, as it learns from more negative choices in most datasets. Interestingly, however, the model predicts all positive answers for 25.63\% instances in PIQA, which is significantly higher than all the other datasets. This is one of the sources of error in PIQA, as the model often predicts both choices as positive, but assigns a higher positive probability to the incorrect choice. We also report the \% of instances for which the correct answer is predicted as positive and all incorrect answers are predicted as negative in the \textbf{Accurate} column. The accuracy is highest in HellaSWAG and lowest in QASC, which co-relates well with the highest performance in HellaSWAG and second lowest performance in QASC across the datasets in \Cref{tab:val-results} and \Cref{tab:test-results}.

\subsection{Error Analysis}
We show some examples of incorrect predictions for the DeBERTa-\model{} model in the CommonsenseQA and PIQA dataset in \Cref{tab:error-analysis}. The erroneously predicted answers in CommonsenseQA are often very close in meaning to the correct answers. Furthermore, the incorrectly predicted answer could also be argued as correct for some instances (second example in \Cref{tab:error-analysis}), as the incorrect choice is also equally plausible. In PIQA however, the model make mistakes where complex scientific and physical world knowledge is required. The incorporation of external knowledge is likely necessary to answer these questions accurately.

\begin{table}[t!]
\centering
\resizebox{\linewidth}{!}{
\begin{tabular}{l|cc|ccc}
\toprule
\multirow{2}{*}{\textbf{Dataset}} & \multicolumn{5}{c}{\textbf{DeBERTa-\model{} Predicted All }}  \\
& \textbf{Neg} & \textbf{Pos} & \textbf{Incor as Neg} & \textbf{Cor as Pos} & \textbf{Accurate}  \\
\toprule
CQA & 17.69 & 0.08 & 70.35 & 76.99 & 52.66 \\
CQA2 & 1.81 & 6.53 & 65.17 & 69.89 & 63.36 \\
QASC & 37.37 & 0.0 & 80.45 & 55.29 & 43.09 \\
SWAG & 13.2 & 0.05 & 86.97 & 85.0 & 73.77 \\
H-SWAG & 15.63 & 0.01 & 94.69 & 83.39 & 79.06 \\
SIQA & 20.93 & 2.61 & 73.69 & 72.36 & 52.76 \\
PIQA & 19.37 & 25.63 & 70.46 & 76.71 & 51.09 \\
CosmosQA & 19.33 & 0.2 & 78.32 & 76.21 & 58.99 \\
CICEROv1 & 22.62 & 0.37 & 80.60 & 71.80 & 57.44 \\
CICEROv2 & 11.26 & 2.64 & 79.40 & 85.71 & 68.14 \\
\bottomrule
\end{tabular}
}
\caption{\footnotesize{DeBERTA-\model{} binary classification results. The \textbf{Neg} and \textbf{Pos} column indicate \% of instances for which all answer choices are predicted as negative or positive. The \textbf{Incor as Neg}, \textbf{Cor as Pos}, and \textbf{Accurate} column indicate \% of instances for which all incorrect answers are predicted as negative, the correct answer is predicted as positive, and all answers are predicted accurately as negative or positive. \textbf{Accurate} is the intersection of \textbf{Incor as Neg} and \textbf{Cor as Pos}.
}}
\label{tab:classifier-analysis}
\end{table}

\begin{table}[h]
\centering
\resizebox{\linewidth}{!}{
\begin{tabular}{p{11cm}}
\toprule
\textbf{Dataset}: CommonsenseQA. \textbf{Question}: Though the thin film seemed fragile, for it's intended purpose it was actually nearly what? \textbf{Correct Answer}: Indestructible. \textbf{Predicted Answer}: Unbreakable. \\
\midrule
\textbf{Dataset}: CommonsenseQA. \textbf{Question}: She was always helping at the senior center, it brought her what? \textbf{Correct Answer}: Happiness. \textbf{Predicted Answer}: Satisfaction. \\
\midrule
\textbf{Dataset}: PIQA. \textbf{Goal}: To discourage house flies from living in your home, \textbf{Correct Answer}: keep basil plants in the kitchen or windows. \textbf{Predicted Answer}: keep lavender plants in the kitchen or window. \\
\midrule
\textbf{Dataset}: PIQA. \textbf{Goal}: To cook perfectly golden pancakes, \textbf{Correct Answer}: keep the temperature low for a longer time. \textbf{Predicted Answer}: keep the temperature high and cook quickly. \\
\bottomrule
\end{tabular}
}
\caption{\footnotesize{Some examples of incorrect predictions in CommonsenseQA and PIQA.}}
\label{tab:error-analysis}
\end{table}

\section{Conclusion}
\looseness=-1 In this paper, we introduced a simple binary classification method as an alternative way to address multi-choice question answering (MCQA) tasks. Through evaluations on ten different MCQA benchmarks, we showed that this simple method generally exceeds the performance of the  score-based method traditionally used in the past. We believe this approach can also be used in the more natural open-ended answer generation setups,  thus providing a ``bridge'' between the MCQA and answer generation frameworks for question answering.

\section{Limitations}

Although the method we introduced is more flexible than the answer scoring approach typically used for MCQA, it still lacks the full flexibility of open-ended question answering and assumes the availability of a candidate answer that it can classify as correct or incorrect.

Additionally, even if our approach outperforms the score-based methods for most of the benchmarks we considered, there are still some datasets (e.g., SIQA, PIQA, CICERO v1), where the score-based method performs best. We leave it for future work to identify a principled approach for selecting the best methodology to use for a given dataset. 

\section*{Acknowledgement}
This research/project is supported by the National Research Foundation, Singapore, and the Ministry of National Development, Singapore under its Cities of Tomorrow R\&D Programme (CoT Award COT-V2-2020-1). Any opinions, findings, and conclusions, or recommendations expressed in this material are those of the author(s) and do not reflect the views of the National Research Foundation, Singapore, and the Ministry of National Development, Singapore. This research is also supported by A*STAR under its RIE 2020 AME programmatic grant RGAST2003 and the Ministry of Education, Singapore, under its AcRF Tier-2 grant (Project no. T2MOE2008, and Grantor reference no. MOET2EP20220-0017). Any opinions, findings, conclusions, or recommendations expressed in this material are those of the author(s) and do not reflect the views of the Ministry of Education, Singapore.

\bibliography{refs}
\bibliographystyle{acl_natbib}

\newpage
\appendix
\section{Experimental Details}
We train all the score-based and classification-based models with the AdamW~\cite{loshchilov2018decoupled} optimizer with a learning rate of {1e-6, 3e-6, 5e-6, 1e-5, 3e-5}. We train all the models for 8 epochs. The best models are chosen based on the results on the validation set. The RoBERTa-Large and DeBERTa-Large models have 355M and 304M parameters, respectively.

\section{Computational Resources}
We use a single Quadro RTX 8000 GPU for our experiments. Training takes between 30 minutes to 8 hours for the different datasets used in the paper. 

\section{Dataset Details}
All datasets used in this paper are in English language. The datasets are available in the corresponding leaderboard websites\footnote{\url{https://leaderboard.allenai.org/}} or through the huggingface datasets hub\footnote{\url{https://huggingface.co/datasets}}.

The number of MCQA instances in the training, validation and test set of the various datasets are shown in \cref{tab:stat}. Some example instances from the datasets are shown in \cref{tab:datasets-appendix}.

\begin{table}[h]
\centering
\resizebox{\linewidth}{!}{
\begin{tabular}{lccc}
\toprule
\textbf{Dataset} & Train & Validation & Test \\
\midrule
Abductive NLI & 169,654 & 1,532 & 3,040\\
Commonsense QA & 9,741 & 1,221 & 1,140 \\
Commonsense QA 2.0 & 9,264 & 2,541 & 2,473 \\
QASC / QASC IR & 8,134 & 926 & 920 \\
SWAG & 73,546 & 20,006 & 20,005 \\
HellaSwag & 39,905 & 10,042 & 10,050 \\
PIQA & 16,113 & 1,838 & 3,446 \\
SIQA & 33,410 & 1,954 & 2,059 \\
CosmosQA & 25,262 & 2,985 & 6,963 \\
CICERO v1 & 27,225 & 9,470 & 9,064 \\
CICERO v2 & 13,496 & 2,806 & 4,150 \\
\bottomrule
\end{tabular}
}
\caption{Number of MCQA instances in the train, validation, and test set for the experimental datasets.}
\label{tab:stat}
\end{table}

\begin{table*}[t]
\small
\centering
\resizebox{\linewidth}{!}{
\begin{tabular}{lcL{10.5cm}}
\toprule
\textbf{Dataset} & \textbf{Task} & \textbf{Instance} \\
\midrule

\multirow{6}{*}{ANLI} & \multirow{6}{*}{Intermediate Event Selection} & 
\textbf{Event 1:} Jenny cleaned her house and went to work, leaving the window just a crack open. \\
& & \textbf{Event 2:} When Jenny returned home she saw that her house was a mess! \\
\cmidrule{3-3} & & \textbf{Choice 1:} \color{Emerald}{A thief broke into the house by pulling open the window.} \\
& & \textbf{Choice 2:} At work, she opened her window and the wind blew her papers everywhere.
\\
\midrule

\multirow{5}{*}{CommonsenseQA} & \multirow{5}{*}{Answer Selection} & 
\textbf{Question:} Where on a river can you hold a cup upright to catch water on a sunny day? \\
\cmidrule{3-3}
& & \textbf{Choice 1:} \color{Emerald}{Waterfall} \hspace{2mm} \color{black}\textbf{Choice 2:} Bridge \hspace{2mm} \textbf{Choice 3:} Valley \\
& & \textbf{Choice 4:} Pebble \hspace{5mm} \textbf{Choice 5:} Mountain \\
\midrule

\multirow{3}{*}{CommonsenseQA 2.0} & \multirow{3}{*}{Answer Selection} & 
\textbf{Question:} The peak of a mountain almost always reaches above the the tree line. \\
\cmidrule{3-3}
& & \textbf{Choice 1:} No \hspace{2mm} \color{black}\textbf{Choice 2:} \color{Emerald}{Yes} \\
\midrule

\multirow{4}{*}{QASC} & \multirow{4}{*}{Answer Selection} & 
\textbf{Question:} Differential heating of air can be harnessed for
what? \\
\cmidrule{3-3}
& & \textbf{Choice 1:} \color{Emerald}{electricity production} \hspace{2mm} \color{black}\textbf{Choice 2:} running and lifting \hspace{2mm} \\
& & \textbf{Choice 3:} animal survival  $\, \ \ \ \dots \quad $ \textbf{Choice 8:} reducing acid rain \\
\midrule

\multirow{6}{*}{SWAG} & \multirow{6}{*}{Ending Prediction} & 
\textbf{Partial Event:} On stage, a woman takes a seat at the piano. She \\
\cmidrule{3-3}
& & \textbf{Ending 1:} sits on a bench as her sister plays with the doll. \\
& & \textbf{Ending 2:} smiles with someone as the music plays. \\
& & \textbf{Ending 3:} is in the crowd, watching the dancers. \\
& & \textbf{Ending 4:} \color{Emerald}{nervously sets her fingers on the keys.} \\
\midrule

\multirow{6}{*}{HellaSwag} & \multirow{6}{*}{Ending Prediction} & 
\textbf{Partial Event:} A woman is outside with a bucket and a dog. The dog is running around trying to avoid a bath. She \\
\cmidrule{3-3}
& & \textbf{Ending 1:} rinses the bucket off with soap and blow dry the dog’s head. \\
& & \textbf{Ending 2:} uses a hose to keep it from getting soapy. \\
& & \textbf{Ending 3:} \color{Emerald}{gets the dog wet, then it runs away again.} \\
& & \textbf{Ending 4:} gets into a bath tub with the dog. \\
\midrule

\multirow{6}{*}{Social IQA} & \multirow{6}{*}{Answer Selection} & 
\textbf{Context:} Alex spilled the food she just prepared all over the floor and it made a huge mess. \\
& & \textbf{Question:} What will Alex want to do next? \\
\cmidrule{3-3}
& & \textbf{Choice 1:} taste the food  \hspace{2mm} \textbf{Choice 2:} \color{Emerald}{mop up} \\
& & \textbf{Choice 3:} run around in the mess\\
\midrule

\multirow{6}{*}{Physical IQA} & \multirow{6}{*}{Solution Selection} & 
\textbf{Goal:} To separate egg whites from the yolk using a water bottle, you should \\
\cmidrule{3-3}
& & \textbf{Solution 1:} \color{Emerald}{Squeeze the water bottle and press it against the yolk. Release, which creates suction and lifts the yolk.} \\
& & \textbf{Solution 2:} Place the water bottle and press it against the yolk. Keep pushing, which creates suction and lifts the yolk.\\
\midrule

\multirow{12}{*}{CosmosQA} & \multirow{12}{*}{Answer Selection} & 
\textbf{Context:} : It's a very humbling experience when you need someone to dress you every morning, tie your shoes, and put your hair up. Every menial task takes an unprecedented amount of effort. It made me appreciate Dan even more. But anyway I shan't dwell on this (I'm not dying after all) and not let it detract from my lovely 5 days with my friends visiting from Jersey \\
& & \textbf{Question:} What's a possible reason the writer needed someone to
dress him every morning? \\
\cmidrule{3-3}
& & \textbf{Chocie 1:} The writer doesn't like putting effort into these tasks. \\
& & \textbf{Chocie 2:} \color{Emerald}{The writer has a physical disability.} \\
& & \textbf{Chocie 3:} The writer is bad at doing his own hair. \\
& & \textbf{Chocie 4:} None of the above choices. \\
\midrule

\multirow{12}{*}{CICERO v2} & \multirow{12}{*}{Answer Selection} & 
\textbf{Dialogue:} \\
& & A: Dad, why are you taping the windows? \\
& & B: Honey, a typhoon is coming. \\
& & A: Really? Wow, I don't have to go to school tomorrow. \\
& & B: Jenny, come and help, we need to prepare more food.  \\
& & A: OK. Dad! I'm coming. \\
& & \textbf{Target:} Jenny, come and help, we need to prepare more food. \\
& & \textbf{Question:} What subsequent event happens or could happen following the target? \\
\cmidrule{3-3}
& & \textbf{Chocie 1:} \color{Emerald}{Jenny and her father stockpile food for the coming days.} \\
& & \textbf{Chocie 2:} Jenny and her father give away all their food. \\
& & \textbf{Chocie 3:} Jenny and her father eat all the food in their refrigerator. \\
& & \textbf{Chocie 4:} Jenny and her father eat all the food in their refrigerator. \\
\bottomrule

\end{tabular}
}
\caption{\footnotesize Illustration of the different datasets used in this work. The answers highlighted in \emerald{green} are the correct answers.}
\label{tab:datasets-appendix}
\end{table*}

\section{Modifications in CICERO}
CICERO v1 and v2 both contain instances with either one or more than one correct answer choices. We make the following modifications in the original datasets to use them in our MCQA setup here, as we assume only one answer is correct for a given MCQA instance:

\vspace{2.05mm}
\noindent \textbf{v1:} We only consider instances which has one annotated correct answer. Each instance in CICERO v1 has five possible answer choices. Thus, the instances selected for our experiments in all the three sets (training, validation, and test split) has one correct answer and four incorrect answers. 

\vspace{2.05mm}
\noindent \textbf{v2:} All instances in CICERO v2 has at-least two correct answers. We consider instances with at-least one incorrect answer and create the MCQA dataset as follows:
\begin{itemize}
    \item If the original CICERO v2 instance has $n$ correct answers, then we will create $n$ MCQA instances from it, each having one of the correct answers and three incorrect answers.
    \item The three incorrect answers will be chosen from the incorrect answers of the original instance. We perform oversampling (some incorrect answers repeated) to create three incorrect answers if there are less than three incorrect answers in the original instance.
\end{itemize}

For example, an instance in CICERO v2 has answer choices: $\{c_1, c_2, i_1, i_2\}$. The correct answers are $\{c_1, c_2\}$ and the incorrect answers are $\{i_1, i_2\}$. We create two MCQA instances from the original instance -- i) with answer choices $\{c_1, i_1, i_2, i_1\}$, and ii) with answer choices $\{c_2, i_1, i_2, i_2\}$.

\end{document}